\documentclass[10pt,conference]{IEEEtran}
\IEEEoverridecommandlockouts
\usepackage{cite}
\usepackage{amsmath,amssymb,amsfonts}
\usepackage{algorithmic}
\usepackage{graphicx}
\usepackage{textcomp}
\usepackage{xcolor}
\def\BibTeX{{\rm B\kern-.05em{\sc i\kern-.025em b}\kern-.08em
    T\kern-.1667em\lower.7ex\hbox{E}\kern-.125emX}}
\begin{document}

\title{Multimodal Learning for Fake News Detection in Short Videos Using Linguistically Verified Data and Heterogeneous Modality Fusion
% {\footnotesize \textsuperscript{*}Note: Sub-titles are not captured for https://ieeexplore.ieee.org  and
}

% \IEEEauthorblockN{1\textsuperscript{st} Shanghong Li}
% \IEEEauthorblockA{
% \textit{Nanyang Technological University}\\
% LISH006665@e.ntu.edu.sg}

\author{
\IEEEauthorblockN{
    Shanghong Li,
    Chiam Wen Qi Ruth,
    Hong Xu,
    }
\IEEEauthorblockA{
\textit{Nanyang Technological University}\\
LISH006665@e.ntu.edu.sg, \{ruth.chiam, xuhong\}@ntu.edu.sg}

\and

\IEEEauthorblockN{Fang Liu$^*$}\thanks{$^*$ Corresponding author.}
\IEEEauthorblockA{
\textit{Singapore University of Social Sciences}\\
liufang@suss.edu.sg}
}

\maketitle

\begin{abstract}
The rapid proliferation of short video platforms has necessitated advanced methods for detecting fake news. This need arises from the widespread influence and ease of sharing misinformation, which can lead to significant societal harm. Current methods often struggle with the dynamic and multimodal nature of short video content. This paper presents HFN, Heterogeneous Fusion Net, a novel multimodal framework that integrates video, audio, and text data to evaluate the authenticity of short video content. HFN introduces a Decision Network that dynamically adjusts modality weights during inference and a Weighted Multi-Modal Feature Fusion module to ensure robust performance even with incomplete data. Additionally, we contribute a comprehensive dataset VESV (VEracity on Short Videos) specifically designed for short video fake news detection. Experiments conducted on the FakeTT and newly collected VESV datasets demonstrate improvements of 2.71\% and 4.14\% in Marco F1 over state-of-the-art methods. This work establishes a robust solution capable of effectively identifying fake news in the complex landscape of short video platforms, paving the way for more reliable and comprehensive approaches in combating misinformation.
\end{abstract}


\section{Introduction}
\label{sec:intro}
Information dissemination channels have evolved significantly over the years, transitioning from traditional paper media and television broadcasts to online video portals. In recent times, short video platforms have rapidly gained popularity, emerging as a new mainstream medium for news dissemination\cite{hendrickx2023newspapers, niu2023building, klug2020jump}. Consequently, the rapid spread of large volumes of information, often with varying credibility, has become a prevalent issue \cite{bu2023combating, sundar2021seeing}. This has created an urgent need for robust methods to detect the authenticity of content shared through short video carriers.
 
Unlike text and images in print media and television news, the decentralized creation of short videos results in news content lacking a fixed paradigm. Short video platforms often offer user-friendly and accessible video editing tools, enabling users to edit content with minimal training\cite{bu2023combating, al2021acceptance}. These characteristics present new challenges for detecting fake news on short video platforms. Most past methods have relied on traditional news detection techniques for text and images, analyzing authenticity by extracting text content from images\cite{hou2019towards, palod2019misleading, papadopoulou2017web}. However, short videos frequently employ rapid visual sequences and music to shape users' first impressions and potentially mislead them, blurring the boundaries between false and real content. In response to these challenges in data types, recent research has increasingly focused on multimodal approaches to content analysis\cite{jagtap2021misinformation, choi2021using, shang2021multimodal}. By establishing cross-modal features, these approaches aim to provide more sensitive and effective methods for detecting false content.

The development of multimodal methods has made significant strides in addressing the complexities of fake news detection\cite{shang2021multimodal, li2022cnn,liu2023covid, ganti2022novel, qi2023fakesv}. By integrating various modalities such as text, images, and audio, multimodal techniques offer a more comprehensive analysis of content. However, the adoption of these methods introduces several key research challenges that this paper aims to address. One primary issue is determining how to balance the contributions of different modalities to ensure that each is appropriately weighted in the overall analysis. In addition, this paper focuses on exploring strategies to solve the problem of missing modalities by dynamically adjusting the importance of each modality and effectively managing missing data to enhance the robustness and accuracy of fake news detection on short video platforms.

This paper proposes a novel multi-modal framework, Heterogeneous Fusion Net (HFN), for fake news detection on short videos, providing predictions of content trustworthiness. To address the issue of modality imbalance, we introduce a Decision Network that dynamically adjusts the weight of each modality during inference. We also introduce a Weighted Multi-Modal Feature Fusion module to work with the Decision Network. HFN mechanism allows the model to adaptively make more informed decisions by leveraging the strengths of the available data. By processing video content clip by clip and incorporating both visual and textual information, the system builds a comprehensive understanding of the content. At the same time, the Decision Network ensures that when one modality is weaker or missing, the model can rely more heavily on the stronger modality, thus improving robustness. In addition to the model, we propose a new dataset, VESV, specifically designed for short video fake news detection, addressing the need for comprehensive multi-modal datasets in this field. The dataset includes diverse short video content with both visual and textual information. VESV is annotated and verified by linguistic experts, ensuring the dataset's accuracy and suitability as a ground truth to test our model.

The remainder of this paper is organized as follows. Section II reviews related works in multi-modality fake news detection. Section III introduces the architecture of our proposed model, detailing the Decision Network and Weighted Multi-modal Fusion Module. Section IV describes our experimental setup, including the new dataset, training process, and evaluation metrics. In Section V, we present the analysis of the results along with the final conclusion.

\section{Method}

\begin{figure*}[ht]
  \centering
  \includegraphics[width=6.5in]{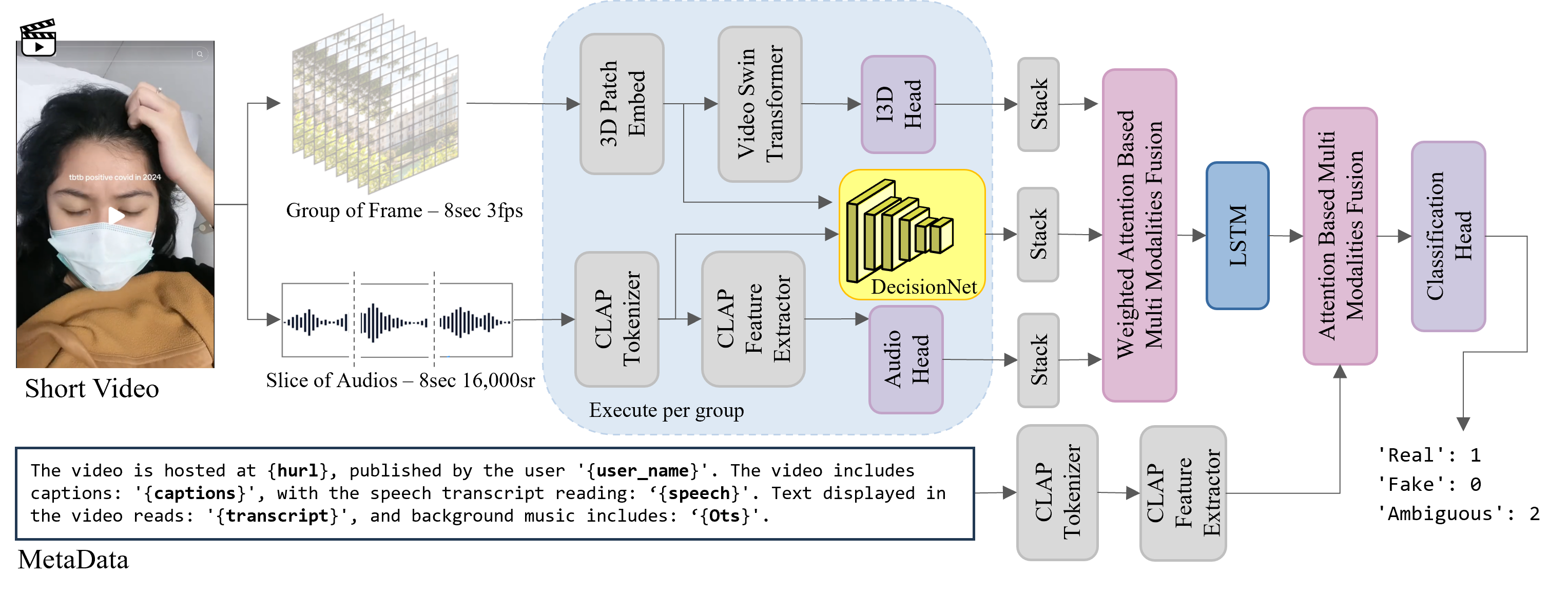}
  \caption{ Overall framework of the proposed multimodal fake news detection model. The video is divided into clips, with each segment of frames and audio input into the blue processing module to extract features. The gray modules represent components that remain static and are not updated during training. Features from the 3D Patch Embed and CLAP Tokenizer are processed through the Video Swin Transformer\cite{liu2022video} and CLAP\cite{wu2023large} Feature Extractor, respectively. Outputs from these modules are fed into the I3D Head and Audio Head, and combined using DecisionNet. The combined features are then processed through LSTM and Attention-Based Multi-Modalities Fusion modules, before reaching the Classification Head for final prediction.}
  \label{fig:Overview}
\end{figure*}

Short videos convey information through multiple channels, including video, audio, and subtitles. Additionally, elements such as usernames, URLs, hashtags, and captions added by the publisher can influence the authenticity of the news. These diverse sources of information correspond to three primary modalities: video, audio, and text. In this section, we provide a detailed description of how the proposed model handles multimodal data for short video fake news detection. The overall framework, as illustrated in Fig.~\ref{fig:Overview}, processes short videos by segmentally extracting and integrating features from the video and audio modalities. It aligns video and audio in the temporal domain by embedding timestamps and subsequently incorporates global text information for comprehensive analysis.

\subsection{Overview}

To begin with, each input video is divided into short clips, denoted as \(\{C_1, C_2, \ldots, C_k\}\), where \(k\) is the total number of clips. The last clip will be padded with zero padding. We set the frame rate of the video to $3fps$ to reduce the cost of calculation. So for each clip \(C_i\), its corresponding frame group could be represented as \(\{F_{i1}, F_{i2}, \ldots, F_{i24}\}\), where \(F_{ij} \in \mathbb{R}^{H \times W \times 3}\). H, W are the height and width of the video frame respectively, and 3 is the RGB channel of the input image. We also cut the audio accordingly into slices with 8 seconds long. The audio sequence is represented as \(S_i\),  where \(S_{i} \in \mathbb{R}^{1 \times Sr }\). $Sr$ is the audio sample rate.

For the visual modality, we utilize a 3D Patch Embedding mechanism to transform each frame group into a set of patches. Specifically, each group \(C_i \in \mathbb{R}^{24 \times H \times W \times 3}\) is divided into group of smaller patches \(P_{i} \in \mathbb{R}^{\frac{24}{2} \times \frac{H}{4} \times \frac{W}{4} \times 96}\). Where 96 is the patch embed dimension. The patch embeddings are then fed into a Video Swin Transformer\cite{liu2022video}, which processes the sequence of embeddings and captures spatiotemporal dependencies. The output of this step is a set of visual feature representations \(V_i \in \mathbb{R}^{\frac{24}{2} \times \frac{H}{32} \times \frac{W}{32} \times d_{Swin}}\), where $d_{Swin}$ is the embed dimension set in Video Swin Transformer backbone.

For the audio modality, the raw audio waveform is first transformed into a Mel-spectrogram using a Mel-filter bank. Then we employ the Contrastive Language-Audio Pretraining (CLAP)\cite{wu2023large} tokenizer, which tokenizes the audio sequence \(S_i\) into a set of audio tokens \(T_i\). These tokens are then passed through the Audio Feature Extractor of CLAP model to generate audio feature representations \(A_i \in \mathbb{R}^{1 \times  d_{CLAP}}\), where $d_{CLAP}$ is the embed dimension set in CLAP backbone.

The extracted visual features \(V_i\) and audio features \(A_i\) are input to the I3D Head and Audio Head, respectively. The structure of these two module are similar, which consist of Pooling layer, Norm layer and MLP layer. Their function is to obtain the features of current frame group \(C_i\) and audio group \(S_i\). The processed features are then combined using the weights \(W_{Vi}\) and  \(W_{Ai}\) output from the DecisionNet, which we will discuss more details in subsection \ref{DN}. The feature and weight of each group will be stacked together to form the feature of whole sequence. Then we proposed a Weighted Attention-based Multi-modal Feature Fusion module to dynamically adjust the influence of each modality according to the weights from DecisionNet, ensuring robust integration even in the presence of missing or weak modalities. More details will be present in subsection \ref{MHA}.

The combined features \(F_i\) are processed through a LSTM\cite{graves2012long} network, capturing temporal dependencies across the video clips. We introduce the global information contained in the text into the model to further enhance the model's ability. Independent tags such as usernames and URLs will be expanded into a sentence as shown in Fig.~\ref{fig:Overview}, making it easier for the pre-trained language model to understand the content. We then use a multimodal fusion module to fuse the global video features with the text features.

Finally, the refined feature representation \(F_{\text{att}}\) is passed through a Classification Head to predict the authenticity of the video content. The classification head consists of a fully connected layer followed by a softmax function, providing the probability distribution over the classes (real or fake).

\subsection{Decision Network}\label{DN}

\begin{figure}
  \centering
  \includegraphics[width=3.2in]{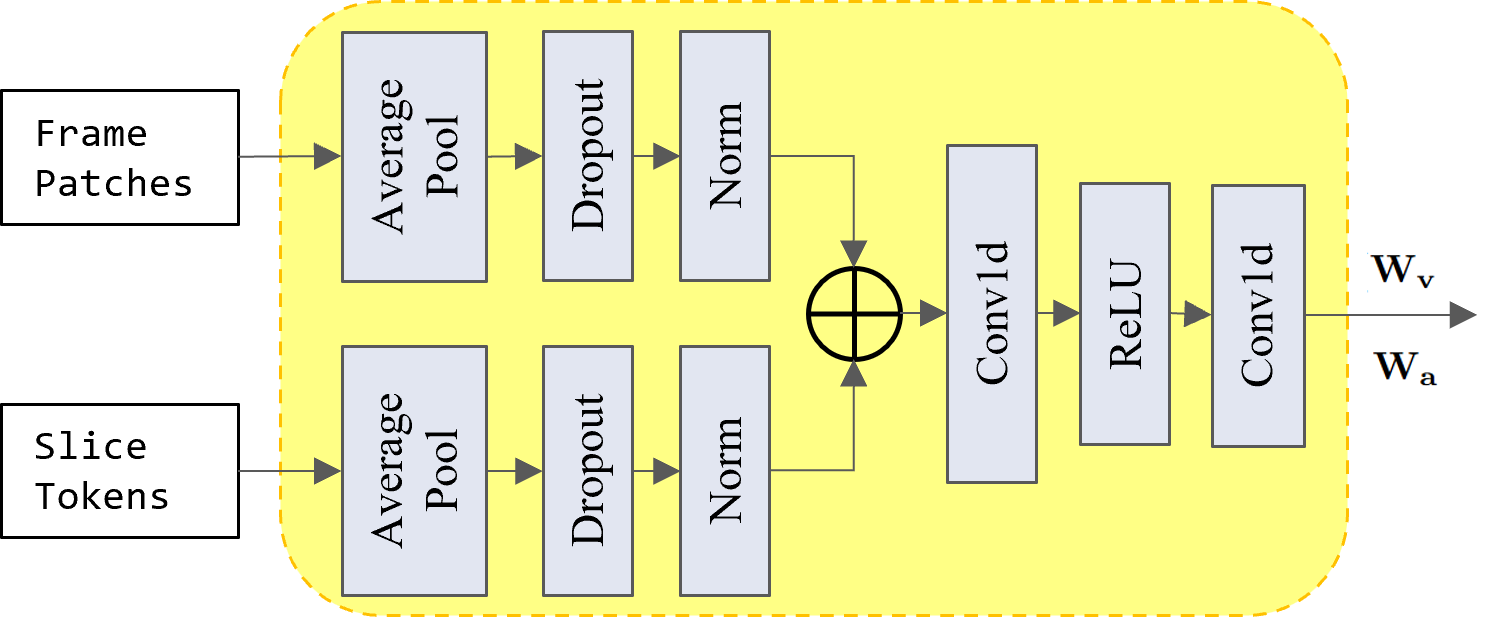}
  \caption{ Details in Decision Net. The outputs of these layers are then summed and passed through a series of Conv1D layers with ReLU activation. The final outputs are denoted as $\mathbf{W_v}$ and $\mathbf{W_a}$, representing the decision weights for video and audio features, respectively.}
  \label{fig:DN}
\end{figure}

The DecisionNet is designed to adaptively weigh the importance of visual and audio features, ensuring robust integration even when certain modalities are weak or missing. The architecture of DecisionNet is depicted in Fig.~\ref{fig:DN}. 

Firstly, the video frame patches \(P_i\) and the audio slice tokens \(T_i\) are separately processed through a series of identical operations. Each modality's feature is subjected to an average pooling layer to reduce its dimensionality and emphasize the most salient aspects. Mathematically, given the visual feature \(V_i\) and audio feature \(T_i\), the average pooled features can be represented as:
\begin{equation}
\hat{P}_i = \text{AvgPool}(P_i), \quad \hat{T}_i = \text{AvgPool}(T_i)
\end{equation}
Next, dropout is applied to each pooled feature to prevent overfitting and improve generalization during training. Subsequently, normalization is performed to stabilize the training process and ensure numerical stability:
\begin{equation}
\bar{P}_i = \text{Norm}(\text{Dropout}(\hat{P}_i)), \quad \bar{T}_i = \text{Norm}(\text{Dropout}(\hat{T}_i))
\end{equation}
The normalized features are then concatenated and passed through a shared convolutional layer layer with ReLU activation to capture interactions between the visual and audio features. Finally, another convolutional layer is applied to produce the decision weights for both modalities
\begin{equation}
H = \text{Conv1D}(\text{ReLU}(\text{Conv1D}(\bar{V}_i \oplus \bar{A}_i)))
\end{equation}
where \(\oplus\) denotes the concatenation operation. 

These weights \(W_{V_i}\) and \(W_{A_i}\) are then used to combine the visual and audio features dynamically, as described in the previous subsection. The features for each group are stacked together to form the feature of the entire sequence, which is then processed by the Weighted Attention-based Multi-modal Feature Fusion module.

\subsection{ Weighted Multi-modal Feature Fusion} \label{MHA}

\begin{figure}
  \centering
  \includegraphics[width=3.2in]{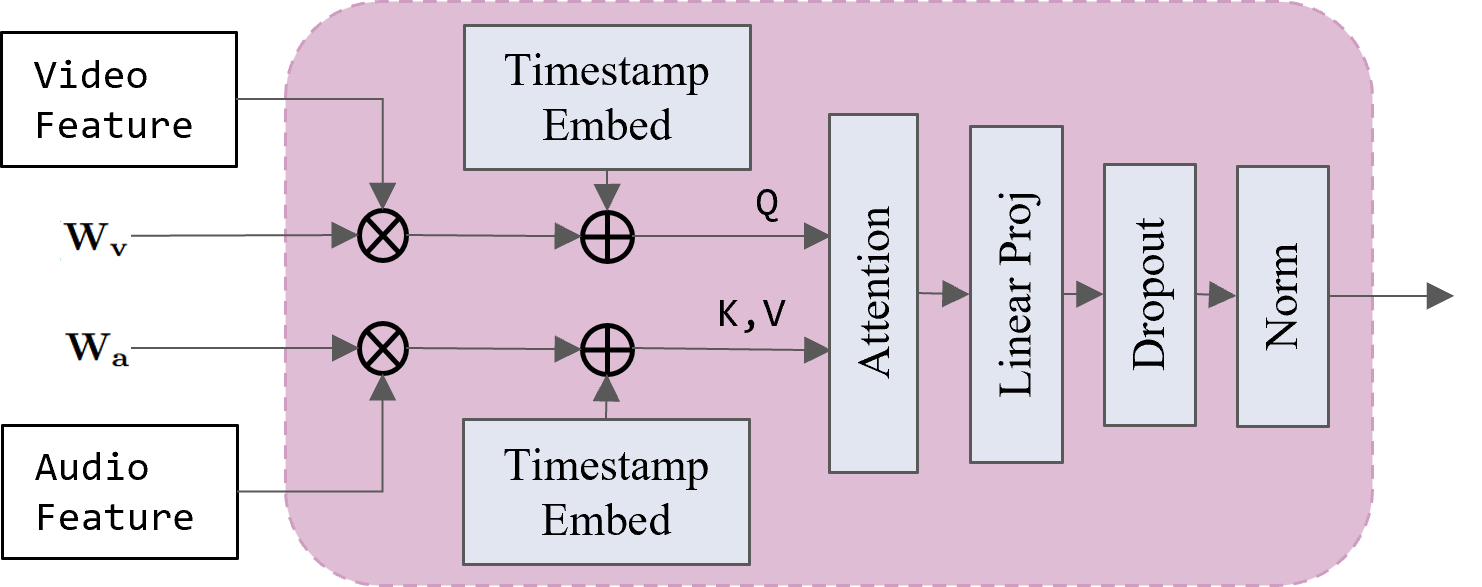}
  \caption{ Details in Weighted Attention Molti-modal Fusion Module. The decision weights $\mathbf{W_v}$ and $\mathbf{W_a}$ from the Decision Network are combined with timestamp embeddings. The combined features are processed through an Attention mechanism followed by Linear Projection, Dropout, and Normalization layers. This fusion enables the integration of temporal information and attention-based feature alignment for the final classification.}
  \label{fig:MHA}
\end{figure}

The Weighted Multi-modal Feature Fusion (WMFF) module integrates the visual and audio features of whole sequence, and adjusts the attention to different modal according to the weights produced by the DecisionNet to ensure effective multimodal information fusion. The detailed architecture of the WMFF module is depicted in Fig.~\ref{fig:MHA}.

The stacked video features and audio features is now represented as \(V\in\mathbb{R}^{L \times d }\) and \(A\in\mathbb{R}^{L \times d }\). The WMFF module takes the stacked weights features \(W_{v}\in\mathbb{R}^{L \times 1 }\) and \(W_{a}\in\mathbb{R}^{L \times 1 }\) from the DecisionNet. After scaling the features using the weights, we get the weighted feature \({W}_{v_i}\) and \(W_{a_i}\). Then the scaled feature will be embedded with timestamp information to maintain temporal consistency. Let \(E_t\) be the timestamp embedding function, the timestamp-embedded features for visual and audio modalities can be represented as:
\begin{equation}
\tilde{W}_{v_i} = W_{v_i} + E_t(t), \quad \tilde{W}_{a_i} = W_{a_i} + E_t(t)
\end{equation}
Then fed the features into a multi-head attention mechanism to capture the inter-dependencies and relationships between the visual and audio features across different time steps. The multi-head attention operation can be described as follows:
\begin{equation}
Q = \tilde{W}_{v_i}, \quad K = V = \tilde{W}_{a_i}
\end{equation}
\begin{equation}
\text{Attention}(Q, K, V) = \text{Softmax}\left(\frac{QK^T}{\sqrt{d_k}}\right)V
\end{equation}
Here, \(d_k\) is the dimensionality of the key vectors. The output of the attention mechanism is then linearly projected, dropout and normalization is applied to prevent overfitting and stabilize the training process:
\begin{equation}
\hat{H} = \text{Norm}(\text{LinearProj}(\text{Attention}(Q, K, V)))
\end{equation}
The normalized output is the final fused feature representation, capturing the integrated multi-modal information effectively. This representation is then used for downstream tasks such as classification or regression, depending on the specific application.

\section{Experiment}
\label{sec:experiment}

\begin{table*}[t]
    \centering
    \caption{Performance comparison between our method and baselines on the FakeTT and new proposed VESV datasets. The bold indicates the best performance. * indicates that the results is directly quoted from the original paper.}
    \begin{tabular}{|l|l|c|c|ccc|ccc|}
        \hline
         &  &  &  & \multicolumn{3}{c|}{Fake (+Ambiguous)} & \multicolumn{3}{c|}{Real} \\
        \cline{5-10}
        Dataset & Method & Accuracy & Macro F1 & Precision & Recall & F1 & Precision & Recall & F1 \\
        \hline
        FakeTT & *HCFC-Medina\cite{serrano2020nlp} & 62.54 & 62.23 & 46.24 & 80.81 & 58.82 & 84.92 & 53.50 & 65.64 \\
        & FANVM\cite{choi2021using} & 71.57 & 70.21 & 55.15 & 75.76 & 63.83 & 85.28 & 69.50 & 76.58 \\
        & *TikTec\cite{shang2021multimodal} & 66.22 & 65.08 & 49.32 & 72.73 & 58.78 & 82.35 & 63.00 & 71.39 \\
        & SVFEND\cite{qi2023fakesv} & 77.14 & 75.63 & 62.33 & 76.33 & 69.57 & 87.91 & 76.33 & 81.69 \\
        & *FakingRecipe\cite{bu2024fakingrecipe} & 79.15 & 77.74 & \textbf{64.75} & 81.82 & 72.18 & 89.74 & 77.83 & 83.30 \\
        \cline{2-10}
        & \textbf{HFN} (Ours) & \textbf{81.73(+3.26\%)} & \textbf{79.85 (+2.71\%)} & 63.56 & \textbf{88.54} & \textbf{73.73} & \textbf{94.12} & \textbf{79.19} & \textbf{85.95} \\
        \hline
        VESV & HCFC-Medina\cite{serrano2020nlp} & 64.34 & 64.28 & 59.14 & 74.19 & 65.82 & 71.54 & 55.86 & 62.74 \\
        & FANVM \cite{choi2021using}& 70.14 & 70.08 & 66.78 & 70.61 & 68.64 & 73.38 & 69.75 & 71.52 \\
        & TikTec\cite{shang2021multimodal} & 68.16 & 68.11 & 62.54 & 77.78 & 69.33 & 75.78 & 59.88 & 66.90 \\
        & SVFEND\cite{qi2023fakesv} & 75.12 & 74.96 & 67.29 & 89.96 & 76.99 & 87.83 & 62.35 & 72.92 \\
        & FakingRecipe\cite{bu2024fakingrecipe} & 77.28 & 77.14 & 69.09 & \textbf{92.11} & 78.96 & \textbf{91.55} & 64.51 & 75.32 \\
        \cline{2-10}
        & \textbf{HFN} (Ours) & \textbf{80.39(+4.02\%)} & \textbf{80.33(+4.14\%)} & \textbf{72.79} & 91.46 & \textbf{80.85} & 90.48 & \textbf{71.31} & \textbf{79.82} \\
        \hline
    \end{tabular}
\label{tab:quantitative} 
    
\end{table*}

This section initiates with an introduction to the foundational configuration of the interactive training and validation strategy. Subsections include a report of accuracy of our proposed algorithm on test dataset, some ablation experiments and the computational analysis.

\subsection{Datasets}

To validate the generalization performance of the proposed model, we conduct experiments on two datasets: FakeTT and our newly collected VESV dataset.

FakeTT\cite{bu2024fakingrecipe} is a publicly available English dataset for fake news detection on short video platforms. The dataset focus on videos from TikTok which related to events reported by the face-checking website Snopes. It includes 1,172 fake news videos and 819 real news from 2019 with video, audio and some text description available.

VESV (VEracity on Short Videos) is our newly collected dataset, comprising 603 TikTok videos (324 real videos, 279 fake videos) published in 2020 or later, covering topics such as Covid-19, climate change, cancer, technology, and some others. The videos range from 10 seconds to 10 minutes in length.Annotation was performed by a team of three annotators following a multi-step process, with the assistance and verification by linguistic experts to ensure the accuracy and reliability of the dataset. Hashtags were standardized and captions that contained only hashtags were reclassified under hashtag annotations. The speech was transcribed using Wav2Vec2\cite{baevski2020wav2vec}, with manual edits for precision. On-screen text, including relevant content from the screenshots, was documented as thoroughly as possible. 
To ensure clarity in defining veracity, we adopt the labeling criteria established in prior benchmark, including SVFEND\cite{qi2023fakesv}, FANVM\cite{choi2021using}, and FakingRecipe\cite{bu2024fakingrecipe}. Following these works, videos are categorized as Fake, Real, or Ambiguous (if it narrates entirely personal stories or unverifiable claims that cannot be cross-checked). Two annotators independently assigned labels to each video. Any discrepancies between the annotators were reviewed and resolved by the first author to ensure consistent classification.

\subsection{Experimental Configuration}\label{Experimental Configuration}

{\bf Training Protocol} The training of the model was conducted using a k-fold training strategy of two dataset. We cropped the video into a size of $224\times 224$. All training was conducted in an end-to-end manner. We used a batch size of 8 and trained the model for 150 epoches. The early stop strategy will be performed when the Marco F1 on validation dataset no longer improve after 30 epoches.

The parameters in the feature extractors, including Video Swin Transformer and CLAP model, were not updated during training. We used the Adam optimizer with $weight\_decay=5e^{-5}$ with the learning rate 0.0001 to optimize the other modules, like DecisionNet and the WMFF. We used the Cosine Annealing strategy to reduce the learning rate, updated by the end of each epoch with the minimum learning rate of 0.000001. We combined the Cross Entropy Loss as the optimization target. The training process was conducted on two NVIDIA RTX6000 Ada GPU. The training process was completed in approximately 12 hours.

{\bf Evaluation Protocol} Evaluation was performed on the FakeTT subset and VESV separately. Since there is no official separation, to ensure fair comparison, we repeatedly trained the model three times during evaluation. Each training will randomly divide the data set into 6 parts, taking one as the testing set, one as the val set, and the rest as the training set. The final result is the average of the three validations. For FakeTT, Each epoch consists of 1,328 training samples, 332 validating samples and 332 testing samples. For our new dataset, each epoch consists of 400 training samples, 100 validating samples and 102 testing samples. We employed average accuracy and Macro F1 score among three classes as the evaluation metrics. This comprehensive evaluation strategy ensured a thorough assessment of the model's effectiveness and robustness.

\subsection{Quantitative analysis}

In this section, we compare our method with current SOTA methods\cite{serrano2020nlp,choi2021using,shang2021multimodal,qi2023fakesv,bu2024fakingrecipe} in terms of accuracy and Marco F1, further analyzing the effectiveness of the proposed models.

\textbf{Comparison with State-of-the-Art} Table~\ref{tab:quantitative} summarizes the quantitative results of our model compared to several state-of-the-art models on the FakeTT and VESV datasets. Our method consistently outperforms all competing methods in both Accuracy and Macro F1 on both datasets, validating its effectiveness in detecting fake news videos. Specifically, on the FakeTT dataset, our model achieves a 2.71\% improvement in Macro F1 score and a 3.26\% improvement in Accuracy over the previous best-performing method, FakingRecipe. On the newly proposed VESV dataset, our model demonstrates even greater improvements, with a 4.14\% increase in Macro F1 score and a 4.02\% increase in Accuracy.

In addition to overall performance, we evaluated our method's effectiveness on both fake (including Ambiguous on VESV) and real samples across the two datasets. Our method achieves the highest F1 scores for both fake and real samples, underscoring its robustness and accuracy. On the FakeTT dataset, while our Precision on fake samples is slightly lower than that of FakingRecipe, our method delivers the best overall performance for real samples. On the VESV dataset, our method exhibits a lower Recall on fake samples and a lower Precision on real samples compared to FakingRecipe. Moreover, we observed a trend where the accuracy of all methods in detecting fake samples on FakeTT is generally lower than for real samples, whereas the opposite trend is observed on VESV. This discrepancy suggests inherent differences in the distribution of the two datasets.

\textbf{Computational Analysis} We analyzed the computational complexity of our Fake News Detection model by examining the parameters and floating point operations per second (FLOPs) of each module. The model architecture is designed to balance accuracy and efficiency. Table~\ref{tab:computation} provides a detailed breakdown of the parameters and FLOPs for each module in the model.

\begin{table}[htbp]
    \caption{Efficiency Comparison with SOTA Methods}
    \centering
    \begin{tabular}{|l|c|c|} 
        \hline
         & \textbf{Params} MB & \textbf{Infer Time} s\\
        \hline
        % HCFC-Medina\cite{serrano2020nlp} & - & - \\
        % FANVM \cite{choi2021using} & - & - \\
        % TikTec \cite{shang2021multimodal} & - & - \\
        SVFEND \cite{qi2023fakesv} & 2,315  & 132\\
        FakingRecipe \cite{bu2024fakingrecipe} & 2,734  & 90\\
        \hline
        \textbf{HFN} (Ours) & 1,834 & 29 \\
        \hline
    \end{tabular}
    \label{tab:computation}
\end{table}

As can be seen from Table~\ref{tab:computation}, compared with previous methods, we have adopted a more lightweight feature extraction model, which greatly reduces the number of parameters of the entire model and the average inference time on both datasets. At the same time, the proposed DecisionNet and WMFF are two lightweight modules. They can improve the overall performance of the model with very little computational overhead.

\subsection{Ablation Study}

To understand the contribution of each component in our model, we conducted an ablation study by systematically removing or modifying specific modules. First, we removed the decision network module and replaced the WMFF (Weighted Multimodal Feature Fusion) module with alternative multimodal feature fusion schemes (Concatenate, Add, Cross-Attention)\cite{zhang2020multimodal} to serve as the baseline models. Next, we set the weights input to the WMFF module to 1, thereby invalidating the weighted aspect of the WMFF module to verify its performance. Finally, we reintroduced the decision network, feeding its output weights into the WMFF module to assess the effectiveness of the decision network module. The results of this study are presented in Table~\ref{tab:ablation}.

\begin{table}[htbp] 
\caption{Ablation Study with input of all modalities on the FakeTT and our dataset, measured by \textbf{F1-score}}
\centering 
\begin{tabular}{|l|c|c|c|c|} 
    \hline
     & \multicolumn{2}{c|}{FakeTT}  &  \multicolumn{2}{c|}{VESV} \\ 
     \hline
    \textbf{Configuration} & Accuracy & Marco F1 &  Accuracy & Marco F1 \\ 
    \hline
    Concatenate & 72.64 & 72.13 & 74.71 & 73.42 \\
    Add & 76.01 & 75.25 & 71.76 & 71.66 \\
    Cross-Attn & 75.69 & 75.43 & 75.36 & 74.40 \\
    \hline
    WMFF without DN & 76.44 & 74.23 & 79.41 & 77.55 \\
    WMFF with DN & \textbf{81.73} & \textbf{79.85} & \textbf{80.39} & \textbf{80.33}\\
    \hline
\end{tabular} 
\label{tab:ablation}
\end{table}

The results in Table~\ref{tab:ablation} indicate that the WMFF module outperforms other multimodal feature fusion schemes. Specifically, the Macro F1 score on the VESV dataset is improved by approximately 4.23\%. Although on the FakeTT dataset, WMFF does not show a significant advantage over the Cross-Attention scheme, the performance is markedly enhanced with the addition of the DecisionNet module. This highlights the critical role of the decision network in our model's superior performance.

We designed ablation experiments to simulate the situation where audio and text are missing in real applications to verify the effectiveness of our method in balancing multimodal performance. The results is presented in Table~\ref{tab:MultiModal}.

\begin{table}[h] 
\caption{Ablation experiments to simulate modality loss} 
\centering 
    \begin{tabular}{|l|c|c|c|c|} 
    \hline 
    \textbf{FakeTT}      & video only  & video audio & video text  & multimodal \\ 
    \hline
    \textbf{FANVM}           & 60.11 & 61.83 & 60.55 & 70.21 \\
    \textbf{TikTec}           & 62.35  & 61.88 & 60.91 & 65.08  \\
    \textbf{SVFEND}         & 70.23  & 69.62 & 75.74  & 75.63 \\
    \textbf{FakingRecipe}   & 71.36  & 71.97 & 72.09 & 77.74  \\
    \hline
    \textbf{HFN} (Ours)    & 75.47 & 77.27 & 77.83 & 79.85 \\
    \hline
    \textbf{VESV}      & video only & video audio & video text & multimodal \\ 
    \hline
    \textbf{FANVM}           & 61.52 & 62.83 & 62.64 & 70.08  \\
    \textbf{TikTec}           & 61.45  & 62.13 & 63.63  & 68.11  \\
    \textbf{SVFEND}           & 71.24 & 71.96 & 72.01 & 74.96 \\
    \textbf{FakingRecipe}   & 72.12 &  69.84 & 70.34 & 77.14 \\
    \hline
    \textbf{HFN} (Ours)   & 75.40 & 77.13 & 77.92 & 80.33 \\
    \hline
\end{tabular} 
\label{tab:MultiModal} 
\end{table}

It can be seen form Table~\ref{tab:MultiModal} that when the audio was missing, our model surpasses all other method, achieving the F1 score of 77.83\% and 77.92\% on each dataset. When the metadata was missing, our model still outperforms others, leading with 77.13\% and 77.27\%. The results highlight the robustness and superior performance of our model, especially in multi-modal scenarios.

It is worth highlighted that our method shows minimal performance degradation when the audio or text modality is absent comparing with other method. Specifically, for the FakeTT dataset, the performance decreases only slightly from 75.47\% to 77.83\% when audio is missing, and from 75.47\% to 77.27\% when text is missing. Similar trends are observed for the VESV dataset. This resilience demonstrates our method's effectiveness in leveraging available information, maintaining high performance even with incomplete multi-modal data, making it a reliable choice for real-world applications.

\section{Conclusion}
\label{sec:conclusion}

This paper presents a comprehensive multi-modal framework for fake news detection on short video platforms, addressing the challenges posed by the diverse and dynamic nature of such content. Our proposed Decision Network and Weighted Multi-Modal Feature Fusion module enable the model to adaptively balance the contributions of different modalities, ensuring robust performance even when certain data types are missing or incomplete. Experimental results on the FakeTT and VESV datasets demonstrate the effectiveness of our approach, showing significant improvements in accuracy and robustness compared to existing methods. Our contributions, including the novel dataset and multi-modal framework, pave the way for more reliable and comprehensive solutions in the fight against misinformation in digital media.

Our study has several limitations. Reliance on the video modality means performance decreases when video data is unavailable. The dataset is relatively small, which may restrict robustness and generalization. To address these issues, future work will focus on improving resilience to missing modalities, expanding the dataset with larger and more diverse collections from multiple platforms, and enhancing the efficiency of training and inference. Optimizing the current clip-based processing pipeline will help reduce computational overhead and improve scalability.

\bibliographystyle{IEEEtranS}
\bibliography{references}

\begin{thebibliography}{10}
\providecommand{\url}[1]{#1}
\csname url@samestyle\endcsname
\providecommand{\newblock}{\relax}
\providecommand{\bibinfo}[2]{#2}
\providecommand{\BIBentrySTDinterwordspacing}{\spaceskip=0pt\relax}
\providecommand{\BIBentryALTinterwordstretchfactor}{4}
\providecommand{\BIBentryALTinterwordspacing}{\spaceskip=\fontdimen2\font plus
\BIBentryALTinterwordstretchfactor\fontdimen3\font minus \fontdimen4\font\relax}
\providecommand{\BIBforeignlanguage}[2]{{%
\expandafter\ifx\csname l@#1\endcsname\relax
\typeout{** WARNING: IEEEtran.bst: No hyphenation pattern has been}%
\typeout{** loaded for the language `#1'. Using the pattern for}%
\typeout{** the default language instead.}%
\else
\language=\csname l@#1\endcsname
\fi
#2}}
\providecommand{\BIBdecl}{\relax}
\BIBdecl

\bibitem{singhal2019spotfake}
S.~Singhal, R.~R. Shah, T.~Chakraborty, P.~Kumaraguru, and S.~Satoh, ``Spotfake: A multi-modal framework for fake news detection,'' in \emph{2019 IEEE fifth international conference on multimedia big data (BigMM)}.\hskip 1em plus 0.5em minus 0.4em\relax IEEE, 2019, pp. 39--47.

\bibitem{zhou2020similarity}
X.~Zhou, J.~Wu, and R.~Zafarani, ``: Similarity-aware multi-modal fake news detection,'' in \emph{Pacific-Asia Conference on knowledge discovery and data mining}.\hskip 1em plus 0.5em minus 0.4em\relax Springer, 2020, pp. 354--367.

\bibitem{singhal2020spotfake+}
S.~Singhal, A.~Kabra, M.~Sharma, R.~R. Shah, T.~Chakraborty, and P.~Kumaraguru, ``Spotfake+: A multimodal framework for fake news detection via transfer learning (student abstract),'' in \emph{Proceedings of the AAAI conference on artificial intelligence}, vol.~34, no.~10, 2020, pp. 13\,915--13\,916.

\bibitem{wang2020fake}
Y.~Wang, S.~Qian, J.~Hu, Q.~Fang, and C.~Xu, ``Fake news detection via knowledge-driven multimodal graph convolutional networks,'' in \emph{Proceedings of the 2020 international conference on multimedia retrieval}, 2020, pp. 540--547.

\bibitem{qian2021hierarchical}
S.~Qian, J.~Wang, J.~Hu, Q.~Fang, and C.~Xu, ``Hierarchical multi-modal contextual attention network for fake news detection,'' in \emph{Proceedings of the 44th international ACM SIGIR conference on research and development in information retrieval}, 2021, pp. 153--162.

\bibitem{qi2021improving}
P.~Qi, J.~Cao, X.~Li, H.~Liu, Q.~Sheng, X.~Mi, Q.~He, Y.~Lv, C.~Guo, and Y.~Yu, ``Improving fake news detection by using an entity-enhanced framework to fuse diverse multimodal clues,'' in \emph{Proceedings of the 29th ACM International Conference on Multimedia}, 2021, pp. 1212--1220.

\bibitem{wu2021multimodal}
Y.~Wu, P.~Zhan, Y.~Zhang, L.~Wang, and Z.~Xu, ``Multimodal fusion with co-attention networks for fake news detection,'' in \emph{Findings of the association for computational linguistics: ACL-IJCNLP 2021}, 2021, pp. 2560--2569.

\bibitem{silva2021embracing}
A.~Silva, L.~Luo, S.~Karunasekera, and C.~Leckie, ``Embracing domain differences in fake news: Cross-domain fake news detection using multi-modal data,'' in \emph{Proceedings of the AAAI conference on artificial intelligence}, vol.~35, no.~1, 2021, pp. 557--565.

\bibitem{palani2022cb}
B.~Palani, S.~Elango, and V.~Viswanathan~K, ``Cb-fake: A multimodal deep learning framework for automatic fake news detection using capsule neural network and bert,'' \emph{Multimedia Tools and Applications}, vol.~81, no.~4, pp. 5587--5620, 2022.

\bibitem{wang2022fmfn}
J.~Wang, H.~Mao, and H.~Li, ``Fmfn: Fine-grained multimodal fusion networks for fake news detection,'' \emph{Applied Sciences}, vol.~12, no.~3, p. 1093, 2022.

\bibitem{tufchi2023comprehensive}
S.~Tufchi, A.~Yadav, and T.~Ahmed, ``A comprehensive survey of multimodal fake news detection techniques: advances, challenges, and opportunities,'' \emph{International Journal of Multimedia Information Retrieval}, vol.~12, no.~2, p.~28, 2023.

\bibitem{hua2023multimodal}
J.~Hua, X.~Cui, X.~Li, K.~Tang, and P.~Zhu, ``Multimodal fake news detection through data augmentation-based contrastive learning,'' \emph{Applied Soft Computing}, vol. 136, p. 110125, 2023.

\bibitem{chen2022cross}
Y.~Chen, D.~Li, P.~Zhang, J.~Sui, Q.~Lv, L.~Tun, and L.~Shang, ``Cross-modal ambiguity learning for multimodal fake news detection,'' in \emph{Proceedings of the ACM web conference 2022}, 2022, pp. 2897--2905.

\bibitem{jing2023multimodal}
J.~Jing, H.~Wu, J.~Sun, X.~Fang, and H.~Zhang, ``Multimodal fake news detection via progressive fusion networks,'' \emph{Information processing \& management}, vol.~60, no.~1, p. 103120, 2023.

\bibitem{qi2023fakesv}
P.~Qi, Y.~Bu, J.~Cao, W.~Ji, R.~Shui, J.~Xiao, D.~Wang, and T.-S. Chua, ``Fakesv: A multimodal benchmark with rich social context for fake news detection on short video platforms,'' in \emph{Proceedings of the AAAI Conference on Artificial Intelligence}, vol.~37, no.~12, 2023, pp. 14\,444--14\,452.

\bibitem{bu2024fakingrecipe}
Y.~Bu, Q.~Sheng, J.~Cao, P.~Qi, D.~Wang, and J.~Li, ``Fakingrecipe: Detecting fake news on short video platforms from the perspective of creative process,'' \emph{arXiv preprint arXiv:2407.16670}, 2024.

\bibitem{fan2016video}
Y.~Fan, X.~Lu, D.~Li, and Y.~Liu, ``Video-based emotion recognition using cnn-rnn and c3d hybrid networks,'' in \emph{Proceedings of the 18th ACM international conference on multimodal interaction}, 2016, pp. 445--450.

\bibitem{xie2017rethinking}
S.~Xie, C.~Sun, J.~Huang, Z.~Tu, and K.~Murphy, ``Rethinking spatiotemporal feature learning for video understanding,'' \emph{arXiv preprint arXiv:1712.04851}, vol.~1, no.~2, p.~5, 2017.

\bibitem{feichtenhofer2019slowfast}
C.~Feichtenhofer, H.~Fan, J.~Malik, and K.~He, ``Slowfast networks for video recognition,'' in \emph{Proceedings of the IEEE/CVF international conference on computer vision}, 2019, pp. 6202--6211.

\bibitem{baevski2020wav2vec}
A.~Baevski, Y.~Zhou, A.~Mohamed, and M.~Auli, ``wav2vec 2.0: A framework for self-supervised learning of speech representations,'' \emph{Advances in neural information processing systems}, vol.~33, pp. 12\,449--12\,460, 2020.

\bibitem{schneider2019wav2vec}
S.~Schneider, A.~Baevski, R.~Collobert, and M.~Auli, ``wav2vec: Unsupervised pre-training for speech recognition,'' \emph{arXiv preprint arXiv:1904.05862}, 2019.

\bibitem{hendrickx2023newspapers}
J.~Hendrickx, ``From newspapers to tiktok: social media journalism as the fourth wave of news production, diffusion and consumption,'' in \emph{Blurring Boundaries of Journalism in Digital Media: New Actors, Models and Practices}.\hskip 1em plus 0.5em minus 0.4em\relax Springer, 2023, pp. 229--246.

\bibitem{niu2023building}
S.~Niu, Z.~Lu, A.~X. Zhang, J.~Cai, C.~F. Griggio, and H.~Heuer, ``Building credibility, trust, and safety on video-sharing platforms,'' in \emph{Extended Abstracts of the 2023 CHI Conference on Human Factors in Computing Systems}, 2023, pp. 1--7.

\bibitem{klug2020jump}
D.~Klug, ``‘jump in and be part of the fun’. how us news providers use and adapt to tiktok,'' in \emph{Midwest Popular Culture Associafion/Midwest American Culture Associafion annual conference (MPCA/ACA)}, 2020.

\bibitem{bu2023combating}
Y.~Bu, Q.~Sheng, J.~Cao, P.~Qi, D.~Wang, and J.~Li, ``Combating online misinformation videos: Characterization, detection, and future directions,'' in \emph{Proceedings of the 31st ACM International Conference on Multimedia}, 2023, pp. 8770--8780.

\bibitem{sundar2021seeing}
S.~S. Sundar, M.~D. Molina, and E.~Cho, ``Seeing is believing: Is video modality more powerful in spreading fake news via online messaging apps?'' \emph{Journal of Computer-Mediated Communication}, vol.~26, no.~6, pp. 301--319, 2021.

\bibitem{al2021acceptance}
R.~Al-Maroof, K.~Ayoubi, K.~Alhumaid, A.~Aburayya, M.~Alshurideh, R.~Alfaisal, and S.~Salloum, ``The acceptance of social media video for knowledge acquisition, sharing and application: A comparative study among youyube users and tiktok users’ for medical purposes,'' \emph{International Journal of Data and Network Science}, vol.~5, no.~3, p. 197, 2021.

\bibitem{ganti2022novel}
D.~Ganti, ``A novel method for detecting misinformation in videos, utilizing reverse image search, semantic analysis, and sentiment comparison of metadata,'' \emph{Utilizing Reverse Image Search, Semantic Analysis, and Sentiment Comparison of Metadata (June 5, 2022)}, 2022.

\bibitem{choi2021using}
H.~Choi and Y.~Ko, ``Using topic modeling and adversarial neural networks for fake news video detection,'' in \emph{Proceedings of the 30th ACM international conference on information \& knowledge management}, 2021, pp. 2950--2954.

\bibitem{shang2021multimodal}
L.~Shang, Z.~Kou, Y.~Zhang, and D.~Wang, ``A multimodal misinformation detector for covid-19 short videos on tiktok,'' in \emph{2021 IEEE international conference on big data (big data)}.\hskip 1em plus 0.5em minus 0.4em\relax IEEE, 2021, pp. 899--908.

\bibitem{sousa2022fighting}
R.~Sousa-Silva, ``Fighting the fake: A forensic linguistic analysis to fake news detection,'' \emph{International Journal for the Semiotics of Law-Revue internationale de S{\'e}miotique juridique}, vol.~35, no.~6, pp. 2409--2433, 2022.

\bibitem{omezi2020proposed}
N.~Omezi and H.~Jahankhani, ``Proposed forensic guidelines for the investigation of fake news,'' \emph{Policing in the Era of AI and Smart Societies}, pp. 231--265, 2020.

\bibitem{urmosne2021phenomena}
G.~{\"U}rm{\"o}sn{\'e}~Simon and E.~Nyitrai, ``The phenomena of epidemic crime, deepfakes, fake news, and the role of forensic linguistics,'' \emph{Inform{\'a}ci{\'o}s T{\'a}rsadalom: T{\'a}rsadalomtudom{\'a}nyi Foly{\'o}irat}, vol. 2021, no.~4, pp. 86--101, 2021.

\bibitem{hou2019towards}
R.~Hou, V.~P{\'e}rez-Rosas, S.~Loeb, and R.~Mihalcea, ``Towards automatic detection of misinformation in online medical videos,'' in \emph{2019 International conference on multimodal interaction}, 2019, pp. 235--243.

\bibitem{palod2019misleading}
P.~Palod, A.~Patwari, S.~Bahety, S.~Bagchi, and P.~Goyal, ``Misleading metadata detection on youtube,'' in \emph{Advances in Information Retrieval: 41st European Conference on IR Research, ECIR 2019, Cologne, Germany, April 14--18, 2019, Proceedings, Part II 41}.\hskip 1em plus 0.5em minus 0.4em\relax Springer, 2019, pp. 140--147.

\bibitem{papadopoulou2017web}
O.~Papadopoulou, M.~Zampoglou, S.~Papadopoulos, and Y.~Kompatsiaris, ``Web video verification using contextual cues,'' in \emph{Proceedings of the 2nd international workshop on multimedia forensics and security}, 2017, pp. 6--10.

\bibitem{serrano2020nlp}
J.~C.~M. Serrano, O.~Papakyriakopoulos, and S.~Hegelich, ``Nlp-based feature extraction for the detection of covid-19 misinformation videos on youtube,'' in \emph{Proceedings of the 1st Workshop on NLP for COVID-19 at ACL 2020}, 2020.

\bibitem{jagtap2021misinformation}
R.~Jagtap, A.~Kumar, R.~Goel, S.~Sharma, R.~Sharma, and C.~P. George, ``Misinformation detection on youtube using video captions,'' \emph{arXiv preprint arXiv:2107.00941}, 2021.

\bibitem{li2022cnn}
X.~Li, X.~Xiao, J.~Li, C.~Hu, J.~Yao, and S.~Li, ``A cnn-based misleading video detection model,'' \emph{Scientific Reports}, vol.~12, no.~1, p. 6092, 2022.

\bibitem{liu2023covid}
F.~Liu, Y.~Yacoob, and A.~Shrivastava, ``Covid-vts: Fact extraction and verification on short video platforms,'' \emph{arXiv preprint arXiv:2302.07919}, 2023.

\bibitem{fernando2017self}
B.~Fernando, H.~Bilen, E.~Gavves, and S.~Gould, ``Self-supervised video representation learning with odd-one-out networks,'' in \emph{Proceedings of the IEEE conference on computer vision and pattern recognition}, 2017, pp. 3636--3645.

\bibitem{ranasinghe2022self}
K.~Ranasinghe, M.~Naseer, S.~Khan, F.~S. Khan, and M.~S. Ryoo, ``Self-supervised video transformer,'' in \emph{Proceedings of the IEEE/CVF Conference on Computer Vision and Pattern Recognition}, 2022, pp. 2874--2884.

\bibitem{jing2020self}
L.~Jing and Y.~Tian, ``Self-supervised visual feature learning with deep neural networks: A survey,'' \emph{IEEE transactions on pattern analysis and machine intelligence}, vol.~43, no.~11, pp. 4037--4058, 2020.

\bibitem{he2020momentum}
K.~He, H.~Fan, Y.~Wu, S.~Xie, and R.~Girshick, ``Momentum contrast for unsupervised visual representation learning,'' in \emph{Proceedings of the IEEE/CVF conference on computer vision and pattern recognition}, 2020, pp. 9729--9738.

\bibitem{grill2020bootstrap}
J.-B. Grill, F.~Strub, F.~Altch{\'e}, C.~Tallec, P.~Richemond, E.~Buchatskaya, C.~Doersch, B.~Avila~Pires, Z.~Guo, M.~Gheshlaghi~Azar \emph{et~al.}, ``Bootstrap your own latent-a new approach to self-supervised learning,'' \emph{Advances in neural information processing systems}, vol.~33, pp. 21\,271--21\,284, 2020.

\bibitem{liu2021swin}
Z.~Liu, Y.~Lin, Y.~Cao, H.~Hu, Y.~Wei, Z.~Zhang, S.~Lin, and B.~Guo, ``Swin transformer: Hierarchical vision transformer using shifted windows,'' in \emph{Proceedings of the IEEE/CVF international conference on computer vision}, 2021, pp. 10\,012--10\,022.

\bibitem{liu2022video}
Z.~Liu, J.~Ning, Y.~Cao, Y.~Wei, Z.~Zhang, S.~Lin, and H.~Hu, ``Video swin transformer,'' in \emph{Proceedings of the IEEE/CVF conference on computer vision and pattern recognition}, 2022, pp. 3202--3211.

\bibitem{chang2022distilhubert}
H.-J. Chang, S.-w. Yang, and H.-y. Lee, ``Distilhubert: Speech representation learning by layer-wise distillation of hidden-unit bert,'' in \emph{ICASSP 2022-2022 IEEE International Conference on Acoustics, Speech and Signal Processing (ICASSP)}.\hskip 1em plus 0.5em minus 0.4em\relax IEEE, 2022, pp. 7087--7091.

\bibitem{wu2023large}
Y.~Wu, K.~Chen, T.~Zhang, Y.~Hui, T.~Berg-Kirkpatrick, and S.~Dubnov, ``Large-scale contrastive language-audio pretraining with feature fusion and keyword-to-caption augmentation,'' in \emph{ICASSP 2023-2023 IEEE International Conference on Acoustics, Speech and Signal Processing (ICASSP)}.\hskip 1em plus 0.5em minus 0.4em\relax IEEE, 2023, pp. 1--5.

\bibitem{graves2012long}
A.~Graves and A.~Graves, ``Long short-term memory,'' \emph{Supervised sequence labelling with recurrent neural networks}, pp. 37--45, 2012.

\bibitem{zhang2020multimodal}
C.~Zhang, Z.~Yang, X.~He, and L.~Deng, ``Multimodal intelligence: Representation learning, information fusion, and applications,'' \emph{IEEE Journal of Selected Topics in Signal Processing}, vol.~14, no.~3, pp. 478--493, 2020.

\end{thebibliography}

\vspace{12pt}

\end{document}